\documentclass[runningheads]{llncs}
\usepackage{graphicx}

\usepackage{tikz}
\usepackage{comment}
\usepackage{amsmath,amssymb} 
\usepackage{color}

\usepackage{multicol}

\usepackage[accsupp]{axessibility}  

\usepackage{enumitem}
\usepackage[pagebackref,breaklinks,colorlinks]{hyperref}
\usepackage{multirow}

\usepackage{marvosym}

\begin{document}

\pagestyle{headings}
\mainmatter

\title{RigNet: Repetitive Image Guided Network for Depth Completion}


\titlerunning{RigNet: Repetitive Image Guided Network for Depth Completion}
\authorrunning{Yan et al.}
\author{Zhiqiang Yan, Kun Wang, Xiang Li, Zhenyu Zhang, \\Jun Li\textsuperscript{\Letter}, and Jian Yang\textsuperscript{\Letter}}
\institute{PCA Lab, Nanjing University of Science and Technology, China\\
\email{\{Yanzq,kunwang,xiang.li.implus,junli,csjyang\}@njust.edu.cn}\\
\email{zhangjesse@foxmail.com}
}

\maketitle
\begin{abstract}
Depth completion deals with the problem of recovering dense depth maps from sparse ones, where color images are often used to facilitate this task. 
Recent approaches mainly focus on image guided learning frameworks to predict dense depth. However, blurry guidance in the image and unclear structure in the depth still impede the performance of the image guided frameworks. To tackle these problems, we explore a repetitive design in our image guided network to gradually and sufficiently recover depth values. Specifically, the repetition is embodied in both the image guidance branch and depth generation branch. In the former branch, we design a repetitive hourglass network to extract discriminative image features of complex environments, which can provide powerful contextual instruction for depth prediction. In the latter branch, we introduce a repetitive guidance module based on dynamic convolution, in which an efficient convolution factorization is proposed to simultaneously reduce its complexity and progressively model high-frequency structures. Extensive experiments show that our method achieves superior or competitive results on KITTI benchmark and NYUv2 dataset.
\end{abstract}

\section{Introduction}
Depth completion, the technique of converting sparse depth measurements to dense ones, has a variety of applications in the computer vision field, such as autonomous driving \cite{hane20173d,cui2019real,wang2021regularizing}, augmented reality \cite{dey2012tablet,song2020channel}, virtual reality \cite{armbruster2008depth}, and 3D scene reconstruction \cite{zhang2019pattern,park2020nonlocal,shen2021distortion,shen2022panoformer}. The success of these applications heavily depends on reliable depth predictions. Recently, multi-modal information from various sensors is involved to help generate dependable depth results, such as color images \cite{ma2018self,chen2019learning}, surface normals \cite{zhang2019pattern,Qiu_2019_CVPR}, confidence maps \cite{2020Confidence,vangansbeke2019}, and even binaural echoes \cite{gao2020visualechoes,parida2021beyond}. Particularly, the latest image guided methods \cite{zhao2021adaptive,liu2021fcfr,hu2020PENet,tang2020learning} principally concentrate on using color images to guide the recovery of dense depth maps, achieving outstanding performance. However, due to the challenging environments and limited depth measurements, it's difficult for existing image guided methods to produce clear image guidance and structure-detailed depth features (see Figs.~\ref{Fig.2} and~\ref{Fig.6}). 
To deal with these issues, in this paper we develop a repetitive design in both the image guidance branch and depth generation branch.

\begin{figure}[t]
\centering
\includegraphics[width=1\columnwidth]{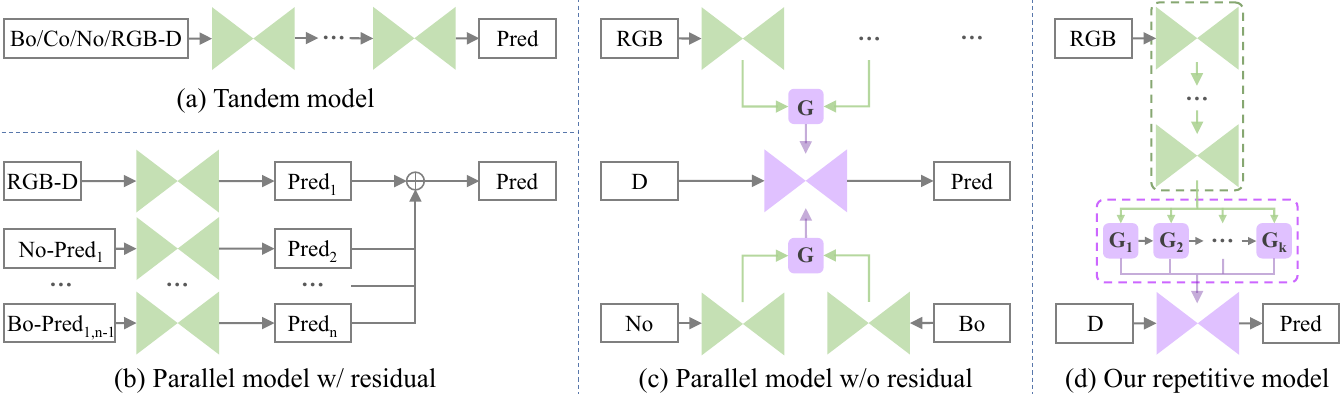}
\caption{To obtain dense depth \textbf{Pred}iction, most existing image guided methods employ tandem models~\cite{ma2018self,Cheng2020CSPN,park2020nonlocal} (a) or parallel models ~\cite{zhao2021adaptive,tang2020learning,liu2021fcfr,hu2020PENet} (b,c) with various inputs (\emph{e.g.}, \textbf{Bo}undary/\textbf{Co}nfidence/\textbf{No}rmal/RGB-D), whilst we propose the repetitive mechanism (d), aiming at providing gradually refined image/depth \textbf{G}uidance.
}\label{model_summary}
\end{figure}

In the image guidance branch: Existing image guided methods are not sufficient to produce very precise details to provide perspicuous image guidance, which limits the content-complete depth recovery. 
For example, the tandem models (Fig.~\ref{model_summary}(a)) tend to only utilize the final layer features of a hourglass unit. The parallel models conduct scarce interaction between multiple hourglass units (Fig.~\ref{model_summary}(b)), or refer to image guidance encoded only by single hourglass unit (Fig.~\ref{model_summary}(c)). Different from them, as shown in Fig.~\ref{model_summary}(d), we present a vertically repetitive hourglass network to make good use of RGB features in multi-scale layers, which contain image semantics with much clearer and richer contexts. 

In the depth generation branch: It is known that gradients near boundaries usually have large mutations, which increase the difficulty of recovering structure-detailed depth for convolution \cite{Uhrig2017THREEDV}. As evidenced in plenty of methods \cite{huang2019indoor,2020Confidence,park2020nonlocal},  the depth values are usually hard to be predicted especially around the region with unclear boundaries. 
To moderate this issue, in this paper we propose a repetitive guidance module based on dynamic convolution \cite{tang2020learning}. It first extracts the high-frequency components by channel-wise and cross-channel convolution factorization, and then repeatedly stacks the guidance unit to progressively produce refined depth. We also design an adaptive fusion mechanism to effectively obtain better depth representations by aggregating depth features of each repetitive unit. However, an obvious drawback of the dynamic convolution is the large GPU memory consumption, especially under the case of our repetitive structure. Hence, we further introduce an efficient module to largely reduce the memory cost but maintain the accuracy.

Benefiting from the repetitive strategy with gradually refined image/depth representations, our method performs better than others, as shown in Figs.~\ref{Fig.4},~\ref{Fig.5} and~\ref{Fig.6}, and reported in Tables~\ref{t2},~\ref{t3},~\ref{t_RHN} and~\ref{t_RG}. In short, our contributions are:
\begin{itemize}
    \item We propose the effective but lightweight repetitive hourglass network, which can extract legible image features of challenging environments to provide clearer guidance for depth recovery.
    \item We present the repetitive guidance module based on dynamic convolution, including an adaptive fusion mechanism and an efficient guidance algorithm, which can gradually learn precise depth representations.
    \item Extensive experimental results demonstrate the effectiveness of our method, which achieves outstanding performances on three datasets.
\end{itemize}

\section{Related Work}
\textbf{Depth only approaches.} 
For the first time in 2017, the work \cite{Uhrig2017THREEDV} proposes sparsity invariant CNNs to deal with sparse depth. Since then, lots of depth completion works \cite{Uhrig2017THREEDV,ku2018defense,2018Deep,2018Sparse,2020Confidence,ma2018self,vangansbeke2019} input depth without using color image. Distinctively, Lu \emph{et al.} \cite{2020FromLu} take sparse depth as the only input with color image being auxiliary supervision when training. However, single-modal based methods are limited without other reference information. As technology quickly develops, plenty of multi-modal information is available, \emph{e.g.}, surface normal and optic flow images, which can significantly facilitate the depth completion task.

\noindent \textbf{Image guided methods.} Existing image guided depth completion methods can be roughly divided into two patterns. One pattern is that various maps are together input into tandem hourglass networks \cite{ma2018self,2018Learning,chen2019learning,Cheng2020CSPN,park2020nonlocal,xu2020deformable}. For example, S2D \cite{ma2018self} directly feeds the concatenation into a simple Unet \cite{ronneberger2015u}. 
CSPN \cite{2018Learning} studies the affinity matrix to refine coarse depth maps with spatial propagation network (SPN). CSPN++ \cite{Cheng2020CSPN} further improves its effectiveness and efficiency by learning adaptive convolutional kernel sizes and the number of iterations for propagation. 
As an extension, NLSPN \cite{park2020nonlocal} presents non-local SPN which focuses on relevant non-local neighbors during propagation. 
Another pattern is using multiple independent branches to model different sensor information and then fuse them at multi-scale stages \cite{vangansbeke2019,2020Denseyang,tang2020learning,li2020multi,liu2021fcfr,hu2020PENet}. For example, PENet \cite{hu2020PENet} employs feature addition to guide depth learning at different stages. 
ACMNet \cite{zhao2021adaptive} chooses graph propagation to capture the observed spatial contexts. GuideNet \cite{tang2020learning} seeks to predict dynamic kernel weights from the guided image and then adaptively extract the depth features. However, these methods still cannot provide very sufficient semantic guidance for the specific depth completion task.

\noindent \textbf{Repetitive learning models.} To extract more accurate and abundant feature representations, many approaches~\cite{ren2015faster,cai2018cascade,liu2020cbnet,qiao2021detectors} propose to repeatedly stack similar components. For example, PANet \cite{liu2018path} adds an extra bottom-up path aggregation which is similar with its former top-down feature pyramid network (FPN). NAS-FPN \cite{ghiasi2019fpn} and BiFPN \cite{tan2020efficientdet} conduct repetitive blocks to sufficiently encode discriminative image semantics for object detection. FCFRNet \cite{liu2021fcfr} argues that the feature extraction in one-stage frameworks is insufficient, and thus proposes a two-stage model, 
which can be regarded as a special case of the repetitive design. On this basis, PENet \cite{hu2020PENet} further improves its performance by utilizing confidence maps and varietal CSPN++.
Different from these methods, in our image branch we first conduct repetitive CNNs units to produce clearer guidance in multi-scale layers. Then in our depth branch we perform repetitive guidance module to generate structure-detailed depth.

 \begin{figure}[t]
  \centering
  \includegraphics[width=0.9\columnwidth]{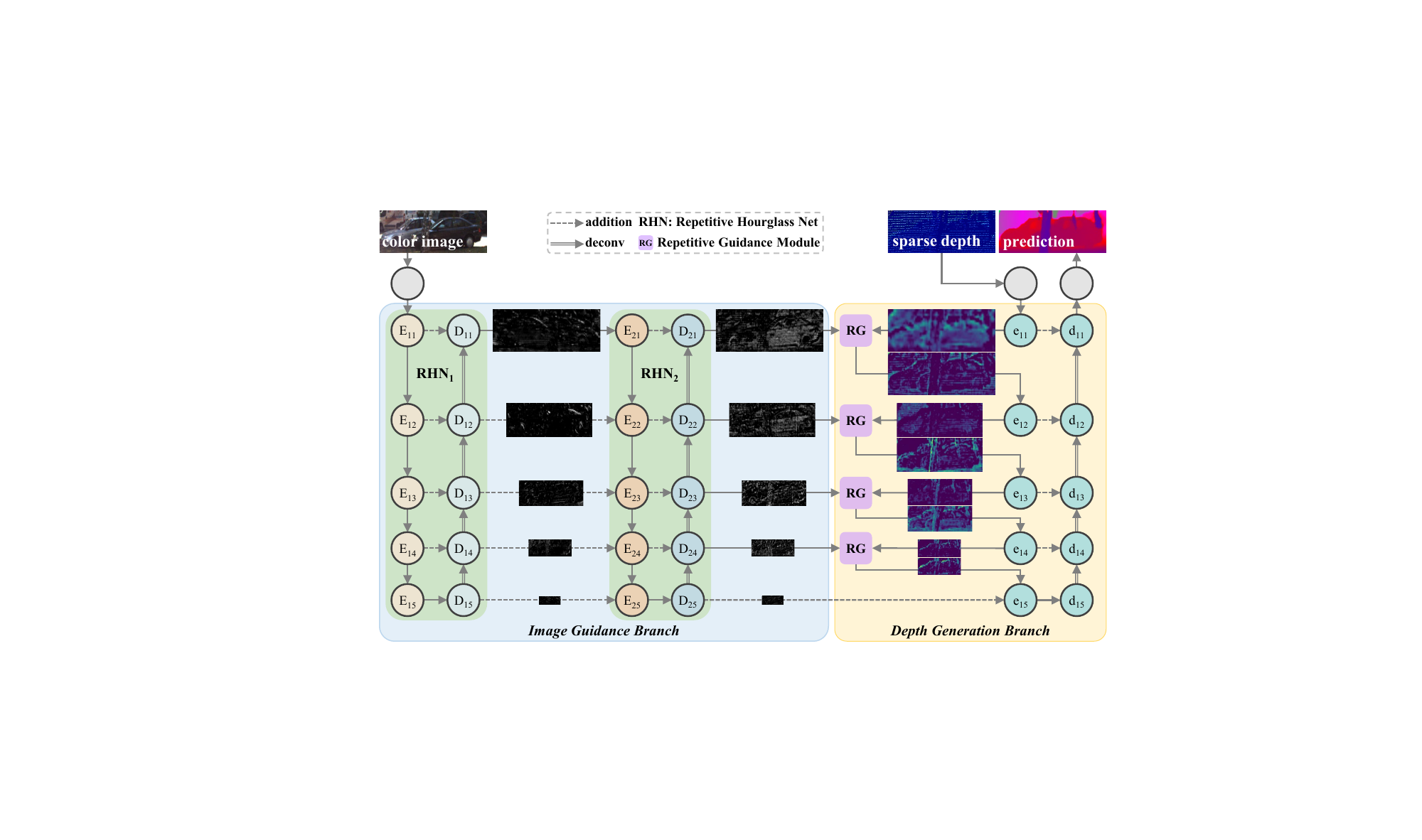}\\
  \caption{Overview of our repetitive image guided network, which contains an image guidance branch and a depth generation branch. The former consists of a repetitive hourglass network (RHN) and the latter has the similar structure as RHN$_1$. In the depth branch, we perform our novel repetitive guidance module (RG, elaborated in Fig.~\ref{Fig.3}) to refine depth. In addition, an efficient guidance algorithm (EG) and an adaptive fusion mechanism (AF) are proposed to further improve the performance of the module.}\label{Fig.2}
\end{figure}

\section{Repetitive Design}
In this section, we first introduce our repetitive hourglass network (RHN), then elaborate the proposed repetitive guidance module (RG), including an efficient guidance algorithm (EG) and an adaptive fusion mechanism (AF).

\subsection{Repetitive Hourglass Network}\label{RHN}
For autonomous driving in challenging environments, it is important to understand the semantics of color images in view of the sparse depth measurement. The problem of blurry image guidance can be mitigated by a powerful feature extractor, which can obtain context-clear semantics. In this paper we present our repetitive hourglass network shown in Fig.~\ref{Fig.2}. RHN$_i$ is a symmetrical hourglass unit like Unet. The original color image is first encoded by a $5\times5$ convolution and then input into RHN$_1$. Next, we repeatedly utilize the similar but lightweight unit, each layer of which consists of two convolutions, to gradually extract high-level semantics. In the encoder of RHN$_i$, $E_{ij}$ takes $E_{i(j-1)}$ and $D_{(i-1)j}$ as input. In the decoder of RHN$_i$, $D_{ij}$ inputs $E_{ij}$ and $D_{i(j+1)}$. When $i>1$, the process is

\begin{equation}\label{e1}
\begin{split}
& {{E}_{ij}}=\left\{ \begin{matrix}
   Conv\left( {{D}_{\left( i-1 \right)j}} \right),\qquad \; \ j=1,  \\
   Conv\left( {{E}_{i\left( j-1 \right)}} \right)+{{D}_{\left( i-1 \right)j}},\, 1<j\le 5,  \\
\end{matrix} \right.\\
& {{D}_{ij}}=\left\{ \begin{matrix}
   Conv\left( {{E}_{i5}} \right),\qquad \qquad \ j=5,  \\
   Deconv\left( {{D}_{i\left( j+1 \right)}} \right)+{{E}_{ij}},\ \ \ 1\le j<5,  \\
\end{matrix} \right.
\end{split}
\end{equation}
where ${Deconv}\left( \cdot \right)$ denotes deconvolution function, and $E_{1j}=Conv(E_{1(j-1)})$.

 \begin{figure}[t]
  \centering
  \includegraphics[width=0.9\columnwidth]{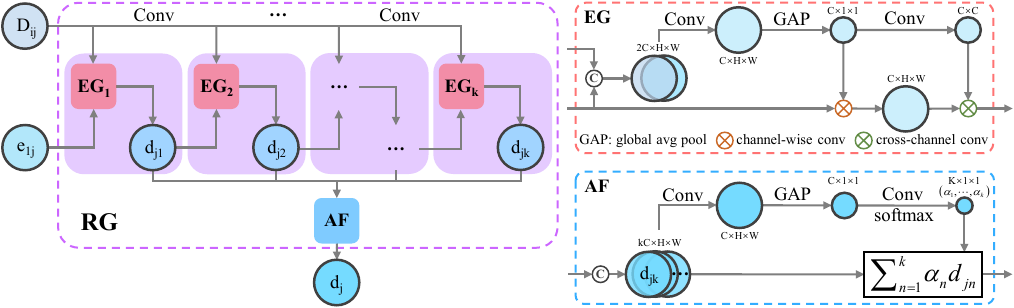}\\
  \caption{Our repetitive guidance (RG) implemented by an efficient guidance algorithm (EG) and an adaptive fusion mechanism (AF). $k$ refers to the repetitive number.}\label{Fig.3}
\end{figure}

\subsection{Repetitive Guidance Module}\label{RG}
Depth in challenging environments is not only extremely sparse but also diverse. Most of the existing methods suffer from unclear structures, especially near the object boundaries. Since gradual refinement is proven effective \cite{Cheng2020CSPN,park2020nonlocal,xu2020deformable} to tackle this issue, we propose our repetitive guidance module to progressively generate dense and structure-detailed depth maps. As illustrated in Fig.~\ref{Fig.2}, our depth generation branch has the same architecture as RHN$_1$. Given the sparse depth input and color image guidance features $D_{ij}$ in the decoder of the last RHN, our depth branch generates final dense predictions. At the stage of the depth branch's encoder, our repetitive guidance module (left of Fig.~\ref{Fig.3}) takes $D_{ij}$ and $e_{1j}$ as input and employs the efficient guidance algorithm (in Sec.~\ref{AG}) to produce refined depth $d_{jk}$ step by step. Then we fuse the refined $d_{jk}$ by our adaptive fusion mechanism (in Sec.~\ref{AF}), obtaining the depth $d_j$, 
\begin{equation}\label{e2}
{{d}_{j}}=RG\left( {{D}_{ij}},{{e}_{1j}} \right),
\end{equation}
where ${RG}\left( \cdot \right)$ refers to the repetitve guidance function.
\vspace{-15pt}

\subsubsection{Efficient Guidance Algorithm}\label{AG}
Suppose the size of inputs $D_{ij}$ and $e_{1j}$ are both $C\times H\times W$. It is easy to figure out the complexity of the dynamic convolution is $O(C\times C\times {{R}^{2}}\times H\times W)$, which generates spatial-variant kernels according to color image features. $R^2$ is the size of the filter kernel window. In fact, $C$, $H$, and $W$ are usually very large, it's thus necessary to reduce the complexity of the dynamic convolution. GuideNet \cite{tang2020learning} proposes channel-wise and cross-channel convolution factorization, whose complexity is $O(C\times {R^2}\times H\times W + C\times C)$. However, our repetitive guidance module employs the convolution factorization many times, where the channel-wise process still needs massive GPU memory consumption, which is $O(C\times {R^2}\times H\times W)$. As a result, inspired by SENet \cite{hu2018squeeze} that captures high-frequency response with channel-wise differentiable operations, we design an efficient guidance unit to simultaneously reduce the complexity of the channel-wise convolution and encode high-frequency components, which is shown in the top right of Fig.~\ref{Fig.3}. Specifically, we first concatenate the image and depth inputs and then conduct a $3\times 3$ convolution. Next, we employ the global average pooling function to generate a $C\times 1\times 1$ feature. At last, we perform pixel-wise dot between the feature and the depth input. The complexity of our channel-wise convolution is only $O(C\times H \times W)$, reduced to ${1}/{{{R}^{2}}}\;$. The process is defined as
\begin{equation}\label{e3}
{{d}_{jk}}=\left\{ \begin{matrix}
   \qquad EG\left( {{D}_{ij}},{{e}_{1j}} \right),\qquad \ \, k=1,  \\
   EG\left( Conv\left( {{D}_{ij}} \right),{{d}_{k-1}} \right),\, k>1,  \\
\end{matrix} \right.
\end{equation}
where ${EG}\left( \cdot \right)$ represents the efficient guidance function. 

Suppose the memory consumptions of the common dynamic convolution, convolution factorization, and our EG are $M_{DC}$, $M_{CF}$, and $M_{EG}$, respectively.

\begin{table}[t]
\centering
\begin{minipage}{0.48\linewidth}
\Large
\resizebox{1.0\columnwidth}{!}{
\begin{tabular}{l}
\hline
$\frac{{{M}_{EG}}}{{{M}_{DC}}}$=$\frac{C\times H\times W+C\times C}{C\times C\times {{R}^{2}}\times H\times W}=\frac{H\times W+C}{C\times H\times W\times {{R}^{2}}}$ \\
$\frac{{{M}_{EG}}}{{{M}_{CF}}}$=$\frac{C\times H\times W+C\times C}{C\times {{R}^{2}}\times H\times W+C\times C}=\frac{H\times W+C}{C+H\times W\times {{R}^{2}}}$ \\\hline
\end{tabular}
}
\caption{Theoretical analysis on GPU memory consumption.}\label{memory_ratio}
\end{minipage}
\begin{minipage}{0.48\linewidth}  
\tiny
\resizebox{1\textwidth}{!}{
\begin{tabular}{l|ccc}
\hline
Method                  & DC     & CF & EG \\\hline
Memory (GB)             & 42.75           & 0.334                 & 0.037  \\
Times (-$/$EG)          & 1155            & 9                     & 1   \\ \hline
\end{tabular}
}
\caption{Numerical analysis on GPU memory consumption.}\label{t1}
\end{minipage}
\end{table}

Table~\ref{memory_ratio} shows the theoretical analysis of GPU memory consumption ratio. Under the setting of the second (4 in total) fusion stage in our depth generation branch, using 4-byte floating precision and taking $C=128$, $H=128$, $W=608$, and $R=3$, as shown in Table~\ref{t1}, the GPU memory of EG is reduced from $42.75$GB to $0.037$GB compared with the common dynamic convolution, nearly 1155 times lower in one fusion stage. Compared to the convolution factorization in GuideNet \cite{tang2020learning}, the memory of EG is reduced from $0.334$GB to $0.037$GB, nearly 9 times lower. Therefore, we can conduct our repetitive strategy easily without worrying much about GPU memory consumption.
\vspace{-15pt}

\subsubsection{Adaptive Fusion Mechanism}\label{AF}
Since many coarse depth features ($d_{j1}$, $\cdots$, $d_{jk}$) are available in our repetitive guidance module, it comes naturally to jointly utilize them to generate refined depth maps, which has been proved effective in various related methods \cite{zhao2017pyramid,lin2017feature,Cheng2020CSPN,song2020channel,park2020nonlocal,hu2020PENet}. Inspired by the selective kernel convolution in SKNet \cite{li2019selective}, we propose the adaptive fusion mechanism to refine depth, which is illustrated in the bottom right of Fig.~\ref{Fig.3}. Specifically, given inputs $(d_{j1}, \cdots, d_{jk})$, we first concatenate them and then perform a $3\times 3$ convolution. Next, the global average pooling is employed to produce a $C\times 1\times 1$ feature map. Then another $3\times 3$ convolution and a softmax function are applied, obtaining $(\alpha_{1},\cdots,\alpha_{k})$,
\begin{equation}\label{e6}
{{\alpha}_{k}}=Soft\left( Conv\left( GAP\left( Conv\left( {{d}_{j1}}|| \cdots ||{{d}_{jk}} \right) \right) \right) \right),
\end{equation}
where $Soft\left( \cdot \right)$ and $||$ refer to softmax function and concatenation. $GAP\left( \cdot \right)$ represents the global average pooling operation.
Finally, we fuse the $k$ coarse depth maps using $\alpha_{k}$ to produce the output $d_j$,
\begin{equation}\label{e7}
{{d}_{j}}=\sum\nolimits_{n=1}^{k}{{{\alpha }_{n}}{{d}_{jn}}}.
\end{equation}
The Eqs.~\ref{e6} and~\ref{e7} can be denoted as
\begin{equation}\label{e8}
{{d}_{j}}=AF\left ( {d}_{j1},{d}_{j2},\cdots,{d}_{jk} \right),
\end{equation}
where $AF\left( \cdot \right)$ represents the adaptive fusion function.

\section{RigNet}
In this section, we describe the network architecture and the loss function for training. The proposed RigNet mainly consists of two parts: (1) an image guidance branch for the generation of hierarchical and clear semantics based on the repetitive hourglass network, and (2) a depth generation branch for structure-detailed depth predictions based on the novel repetitive guidance module with an efficient guidance algorithm and an adaptive fusion mechanism. 

\subsection{Network Architecture}
Fig.~\ref{Fig.2} shows the overview of our network. In our image guidance branch, the RHN$_1$ encoder-decoder unit is built upon residual networks \cite{He2016Deep}. In addition, we adopt the common connection strategy \cite{ronneberger2015u,chen2019learning} to simultaneously utilize low-level and high-level features. RHN$_i$ ($i>1$) has the similar but lightweight architecture with RHN$_1$, which is used to extract clearer image guidance semantics \cite{zeiler2014visualizing}. 

The depth generation branch has the same structure as RHN$_1$. In this branch, we perform repetitive guidance module based on dynamic convolution to gradually produce structure-detailed depth features at multiple stages, which is shown in Fig.~\ref{Fig.3} and described in Sec.~\ref{RG}. 

\subsection{Loss Function}
During training, we adopt the mean squared error (MSE) to compute the loss, 
which is defined as
\begin{equation}\label{e9}
\begin{split}
\mathcal{L}=\frac{1}{m}\sum\limits_{q\in {{Q}_{v}}}{\left\| GT_{q}-{P}_{q} \right\|}^{2},
\end{split}
\end{equation}
where $GT$ and $P$ refer to ground truth depth and predicted depth respectively. $Q_v$ represents the set of valid pixels in $GT$, $m$ is the number of the valid pixels.

\section{Experiments}
In this section, we first introduce the related datasets, metrics, and implementation details. Then, we carry out extensive experiments to evaluate the performance of our method against other state-of-the-art approaches. Finally, a number of ablation studies are employed to verify the effectiveness of our method.

\subsection{Datasets and Metrics}
\textbf{KITTI Depth Completion Dataset} \cite{Uhrig2017THREEDV} is a large autonomous driving real-world benchmark from a driving vehicle. It consists of 86,898 ground truth annotations with aligned sparse LiDAR maps and color images for training, 7,000 frames for validation, and another 1,000 frames for testing. The official 1,000 validation images are used during training while the remained images are ignored. 
Since there are rare LiDAR points at the top of depth maps, the input images are bottom center cropped \cite{vangansbeke2019,tang2020learning,zhao2021adaptive,liu2021fcfr} to $1216\times 256$.

\noindent \textbf{Virtual KITTI Dataset} \cite{gaidon2016virtual} is a
synthetic dataset cloned from the
real world KITTI video sequences. In addition, it also produces color images under various lighting (\emph{e.g.}, sunset, morning) and weather (\emph{e.g.}, rain, fog) conditions. Following GuideNet \cite{tang2020learning}, we use the masks generated from sparse depths of KITTI dataset to obtain sparse samples. Such strategy makes it closed to real-world situation for the distribution of sparse depths.
Sequences of 0001, 0002, 0006, and 0018 are used for training, 0020 with various lighting and weather conditions is used for testing. It contributes to 1,289 frames for fine-tuning and 837 frames for evaluating each condition.

\noindent \textbf{NYUv2 Dataset} \cite{silberman2012indoor} is comprised of video sequences from a variety of indoor scenes as recorded by both the color and depth cameras from the Microsoft Kinect. Paired color images and depth maps in 464 indoor scenes are commonly used. Following previous depth completion methods \cite{ma2018self,chen2019learning,Qiu_2019_CVPR,park2020nonlocal,tang2020learning}, we train our model on 50K images from the official training split, and test on the 654 images from the official labeled test set. Each image is downsized to $320\times 240$, and then $304\times 228$ center-cropping is applied. As the input resolution of our network must be a multiple of 32, we further pad the images to $320\times 256$, but evaluate only at the valid region of size $304\times 228$ to keep fair comparison with other methods.

\noindent \textbf{Metrics.} For the outdoor KITTI depth completion dataset, following the KITTI benchmark and existing methods \cite{park2020nonlocal,tang2020learning,liu2021fcfr,hu2020PENet}, we use four standard metrics for evaluation, including RMSE, MAE, iRMSE, and iMAE. For the indoor NYUv2 dataset, following previous works \cite{chen2019learning,Qiu_2019_CVPR,park2020nonlocal,tang2020learning,liu2021fcfr}, three metrics are selected for evaluation, including RMSE, REL, and ${{\delta }_{i}}$ ($i=1.25, 1.25^2, 1.25^3$).

\subsection{Implementation Details}
The model is particularly trained with 4 TITAN RTX GPUs. We train it for 20 epochs with the loss defined in Eq.~\ref{e9}. We use ADAM \cite{Kingma2014Adam} as the optimizer with the momentum of $\beta_{1}=0.9$, $\beta_{2}=0.999$, a starting learning rate of $1 \times {10}^{-3}$, and weight decay of $1 \times {10}^{-6}$. The learning rate drops by half every 5 epochs. The synchronized cross-GPU batch normalization \cite{ioffe2015batch,zhang2018context} is used when training.

\begin{table}[t]
\centering
\begin{minipage}{0.471\linewidth}
\resizebox{1\columnwidth}{!}{
\begin{tabular}{l|c|cccc}
\hline
\multirow{2}{*}{Method}             & RMSE      & MAE      & iRMSE   & iMAE  \\ 
                                   & mm        & mm       & 1/km    & 1/km \\\hline\hline
CSPN \cite{2018Learning}          & 1019.64   & 279.46   & 2.93    & 1.15  \\
BDBF \cite{Qu_2021_ICCV}            & 900.38    & 216.44   & 2.37    & 0.89  \\
TWISE \cite{imran2021depth}         & 840.20    & \textbf{195.58}    & 2.08 & \textbf{0.82}  \\
NConv \cite{2020Confidence}        & 829.98    & 233.26   & 2.60    & 1.03  \\
S2D \cite{ma2018self}          & 814.73    & 249.95   & 2.80    & 1.21 \\
Fusion \cite{vangansbeke2019}     & 772.87    & 215.02   & 2.19    & 0.93 \\
DLiDAR \cite{Qiu_2019_CVPR}       & 758.38    & 226.50   & 2.56    & 1.15  \\
Zhu \cite{zhu2021robust}           & 751.59    & 198.09   & 1.98    & 0.85 \\
ACMNet \cite{zhao2021adaptive}     & 744.91    & 206.09   & 2.08    & 0.90 \\
CSPN++ \cite{Cheng2020CSPN}        & 743.69    & 209.28   & 2.07    & 0.90  \\
NLSPN \cite{park2020nonlocal}      & 741.68    & 199.59   & \textbf{1.99}  & 0.84\\
GuideNet \cite{tang2020learning}    & 736.24    & 218.83   & 2.25    & 0.99  \\
FCFRNet \cite{liu2021fcfr}          & 735.81    & 217.15   & 2.20    & 0.98  \\
PENet \cite{hu2020PENet}           & 730.08    & 210.55   & 2.17    & 0.94  \\ \hline
RigNet (ours)                       & \textbf{712.66}  & 203.25  & 2.08  & 0.90  \\ \hline
\end{tabular}
}
\caption{Quantitative comparisons on \href{http://www.cvlibs.net/datasets/kitti/eval_depth.php?benchmark=depth_completion}{KITTI depth completion benchmark}.
}\label{t2}
\end{minipage}
\begin{minipage}{0.484\linewidth}  
\resizebox{1.0\textwidth}{!}{
\begin{tabular}{l|ccccc}
\hline
\multirow{2}{*}{Method}      & RMSE      & REL     & \multirow{2}{*}{${\delta }_{1.25}$} & \multirow{2}{*}{${\delta }_{{1.25}^{2}}$} & \multirow{2}{*}{${\delta }_{{1.25}^{3}}$} \\ 
                             & m         & m       & & & \\\hline\hline
Bilateral \cite{silberman2012indoor}     & 0.479 & 0.084 & 92.4 & 97.6 & 98.9\\
Zhang \cite{zhang2018deep} & 0.228 & 0.042 & 97.1 & 99.3 & 99.7 \\
S2D\_18 \cite{ma2018sparse}      & 0.230    & 0.044   & 97.1  & 99.4  & 99.8 \\
DCoeff \cite{imran2019depth} & 0.118    & 0.013   & 99.4  & \textbf{99.9}  & -   \\
CSPN \cite{2018Learning}     & 0.117    & 0.016   & 99.2  & \textbf{99.9}  & \textbf{100.0} \\
CSPN++ \cite{Cheng2020CSPN}      & 0.116    & -       & -     & -     & -    \\
DLiDAR \cite{Qiu_2019_CVPR}   & 0.115    & 0.022   & 99.3  & \textbf{99.9}  & \textbf{100.0}\\
Xu \emph{et al.} \cite{Xu2019Depth}   & 0.112    & 0.018   & 99.5  & \textbf{99.9}  & \textbf{100.0}  \\
FCFRNet \cite{liu2021fcfr}       & 0.106    & 0.015   & 99.5  & \textbf{99.9}  & \textbf{100.0} \\
ACMNet \cite{zhao2021adaptive}   & 0.105    & 0.015   & 99.4  & \textbf{99.9}  & \textbf{100.0}\\
PRNet \cite{lee2021depth}        & 0.104    & 0.014   & 99.4  & \textbf{99.9}  & \textbf{100.0} \\
GuideNet \cite{tang2020learning} & 0.101    & 0.015   & 99.5  & \textbf{99.9}  & \textbf{100.0} \\
TWISE \cite{imran2021depth}      & 0.097    & 0.013   & \textbf{99.6}  & \textbf{99.9} & \textbf{100.0} \\ 
NLSPN \cite{park2020nonlocal}    & 0.092    & \textbf{0.012}   & \textbf{99.6}  & \textbf{99.9}  & \textbf{100.0} \\ \hline
RigNet (ours)                    & \textbf{0.090}    & 0.013   & \textbf{99.6}  & \textbf{99.9}  & \textbf{100.0}\\ \hline
\end{tabular}
}
\caption{Quantitative comparisons on NYUv2 dataset.}\label{t3}
\end{minipage}
\end{table}

\subsection{Evaluation on KITTI Dataset}
\begin{figure}[t]
  \centering
  \includegraphics[width=0.98\columnwidth]{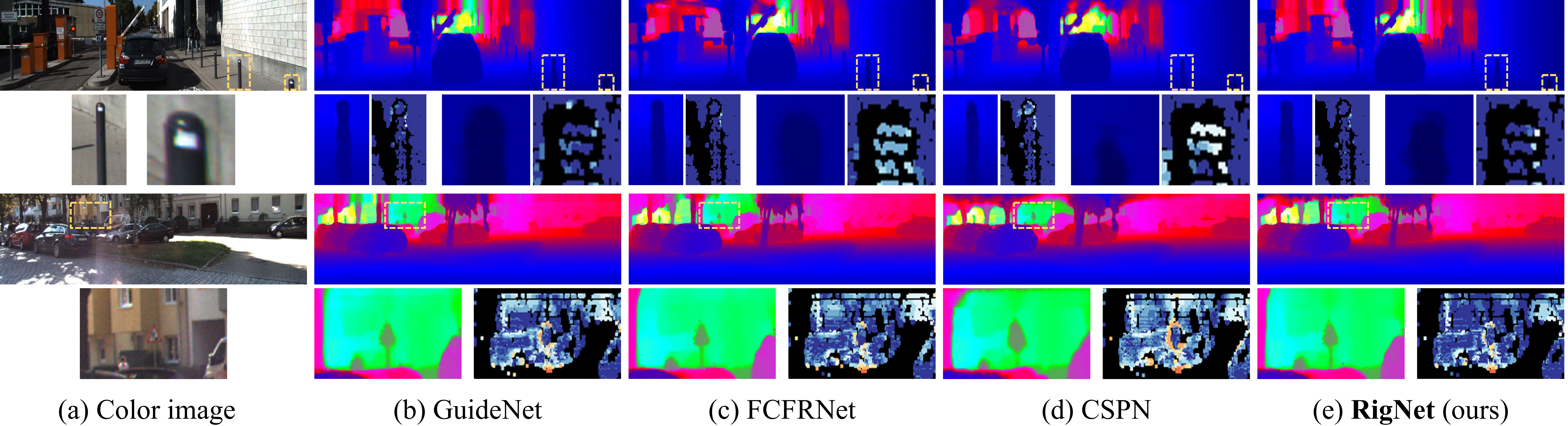}\\
  \caption{Qualitative results on KITTI depth completion test set, including (b) GuideNet \cite{tang2020learning}, (c) FCFRNet \cite{liu2021fcfr}, and (d) CSPN \cite{2018Learning}. Given sparse depth maps and the aligned color images (1st column), depth completion models output dense depth predictions (\emph{e.g.}, 2nd column). We provide \textbf{error maps} borrowed from the KITTI leaderboard for detailed discrimination. Warmer color in error maps refer to higher error.}\label{Fig.4}
\end{figure}

Table~\ref{t2} shows the quantitative results on KITTI benchmark, whose dominant evaluation metric is the RMSE. Our RigNet ranks 1st among publicly published papers when submitting, outperforming the 2nd with significant $17.42mm$ improvement while the errors of other methods are very closed. Here, the performance of our RigNet is also better than those approaches that employ additional dataset, \emph{e.g.}, DLiDAR \cite{Qiu_2019_CVPR} utilizes CARLA \cite{dosovitskiy2017carla} to predict surface normals for better depth predictions. Qualitative comparisons with several state-of-the-art works are shown in Fig.~\ref{Fig.4}. While all methods provide visually good results in general, our estimated depth maps possess more details and more accurate object boundaries. The corresponding error maps can offer supports more clearly. For example, among the marked iron pillars in the first row of Fig.~\ref{Fig.4}, the error of our prediction is significantly lower than the others.

\subsection{Evaluation on NYUv2 Dataset}
\begin{figure}[t]
  \centering
  \includegraphics[width=0.98\columnwidth]{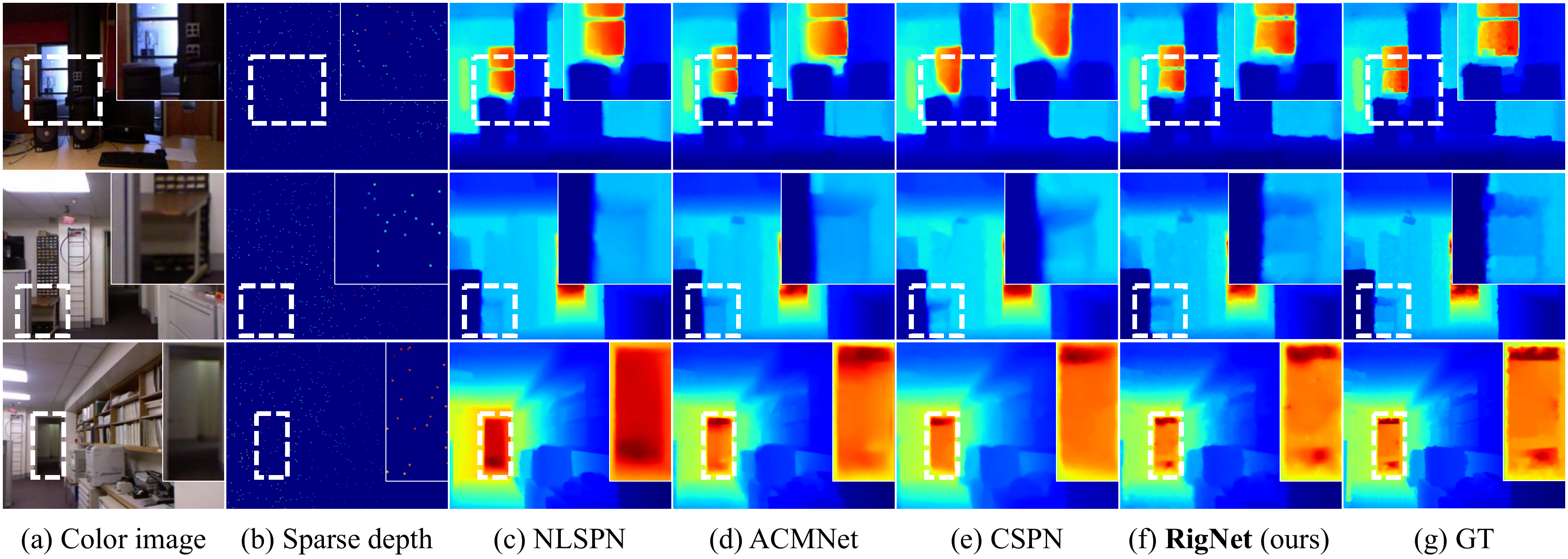}\\
  \caption{Qualitative results on NYUv2 test set. From left to right: (a) color image, (b) sparse depth, (c) NLSPN \cite{park2020nonlocal}, (d) ACMNet \cite{zhao2021adaptive}, (e) CSPN \cite{2018Learning}, (f) our RigNet, and (g) ground truth. We present the results of these four methods under 500 samples. The circled rectangle areas show the recovery of object details.}\label{Fig.5}
\end{figure}

To verify the performance of proposed method on indoor scenes, following existing approaches \cite{Cheng2020CSPN,park2020nonlocal,tang2020learning,liu2021fcfr}, we train our repetitive image guided network on the NYUv2 dataset \cite{silberman2012indoor} with the setting 500 sparse samples. As illustrated in Table~\ref{t3}, our model achieves the best performance among all traditional and latest approaches without using additional datasets, which proves that our network possesses stronger generalization capability. Fig.~\ref{Fig.5} demonstrates the qualitative visualization results. Obviously, compared with those state-of-the-art methods, our RigNet can recover more detailed structures with lower errors at most pixels, including sharper boundaries and more complete object shapes. For example, among the marked doors in the last row of Fig.~\ref{Fig.5}, our prediction is very close to the ground truth, while others either have large errors in the whole regions or have blurry shapes on specific objects.

\begin{table}[t]
\centering
\renewcommand\arraystretch{1.2}
\resizebox{1\textwidth}{!}{
\begin{tabular}{l|ccccc|ccc|ccc|cccc}
\hline
\multirow{2}{*}{Method} 
& \multicolumn{5}{c|}{Deeper} & \multicolumn{3}{c|}{More} & \multicolumn{3}{c|}{Deeper-More} & \multicolumn{4}{c}{Our parallel RHN}   \\ \cline{2-16} 
& 10-1  & \textcolor{blue}{18-1}   & 26-1  & 34-1  & 50-1      
& 18-2  & 18-3  & 18-4     
& 34-2  & 34-3  & 50-2    
& 10-2  & 10-3  & 18-2  & 18-3  \\ \hline \hline
Parameter (M) & 59 & \textcolor{blue}{63}  & 65  & 71  &  84  & 72  & 81 & 91  & 89 & 107 &  104  & 60   &  61   & 64  & 65       \\
Model size (M) & 224 & \textcolor{blue}{239} & 246 & 273 & 317 & 274 & 309 & 344 & 339 & 407 & 398 & 228 & 232 & 242 & 246 \\
RMSE (mm)    & 822  & \textcolor{blue}{779}   & 780   & 778     & 777   & 802     &  816    & 811 & 807  & 801 & 800 & 803 & 798 & 772     & \textbf{769}    \\ \hline
\end{tabular}
}
\caption{Ablation studies of RHN on KITTI validation set. \textcolor{blue}{18-1} denotes that we use 1 ResNet-18 as backbone, which is also the baseline. `Deeper'/`More' denotes that we conduct single\&deeper/multiple\&tandem hourglass units as backbone. Note that each layer of RHN$_{2,3}$ only contains two convolutions while the RHN$_{1}$ employs ResNet.}
\label{t_RHN}
\end{table}

\subsection{Ablation Studies}
Here we employ extensive experiments to verify the effectiveness of each proposed component, including the repetitive hourglass network (RHN-Table~\ref{t_RHN}) and the repetitive guidance module (RG-Table~\ref{t_RG}), which consists of the efficient guidance algorithm (EG) and the adaptive fusion mechanism (AF). \textbf{Note that} the batch size is set to 8 when computing the GPU memory consumption.

\noindent \textbf{(1) Effect of Repetitive Hourglass Network.}

The state-of-the-art baseline GuideNet \cite{tang2020learning} employs 1 ResNet-18 as backbone and guided convolution G$_1$ to predict dense depth. To validate the effect of our RHN, we explore the backbone design of the image guidance branch for the specific depth completion task from four aspects, which are illustrated in Table~\ref{t_RHN}.

\begin{figure*}[t]
  \centering
  \includegraphics[width=1\columnwidth]{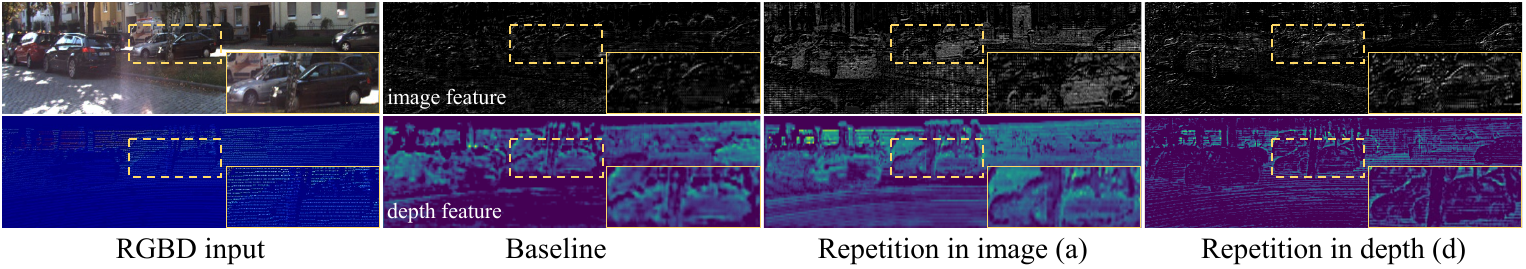}\\
  \caption{Visual comparisons of intermediate features of the baseline and our repetition.}\label{Fig.6}
\end{figure*}

\textbf{(i) Deeper single backbone vs. RHN.} The second column of Table~\ref{t_RHN} shows that, when replacing the single ResNet-10 with ResNet-18, the error is reduced by 43$mm$. However, when deepening the baseline from 18 to 26/34/50, the errors have barely changed, which indicate that simply increasing the network depth of image guidance branch cannot deal well with the specific depth completion task. Differently, with little sacrifice of parameters ($\sim$2 M), our RHN-10-3 and RHN-18-3 are 24$mm$ and 10$mm$ superior to Deeper-10-1 and Deeper-18-1, respectively. Fig.~\ref{Fig.6} shows that the image feature of our parallel RHN-18-3 has much clearer and richer contexts than that of the baseline Deeper-18-1.

\textbf{(ii) More tandem backbones vs. RHN.} As shown in the third column of Table~\ref{t_RHN}, we stack the hourglass unit in series. The models of More-18-2, More-18-3, and More-18-4 have worse performances than the baseline Deeper-18-1. It turns out that the combination of tandem hourglass units is not sufficient to provide clearer image semantic guidance for the depth recovery. In contrast, our parallel RHN achieves better results with fewer parameters and smaller model sizes. These facts give strong evidence that the parallel repetitive design in image guidance branch is effective for the depth completion task.

\textbf{(iii) Deeper-More backbones vs. RHN.} As illustrated in the fourth column of Table~\ref{t_RHN}, deeper hourglass units are deployed in serial way. We can see that the Deeper-More combinations are also not very effective, since the errors of them are higher than the baseline while RHN's error is 10$mm$ lower. It verifies again the effectiveness of the lightweight RHN design. 

\noindent \textbf{(2) Effect of Repetitive Guidance Module.}

\textbf{(i) Efficient guidance.} Note that we directly output the features in EG$_{3}$ when not employing AF. Tables~\ref{memory_ratio} and~\ref{t1} have provided quantitative analysis in theory for EG design. Based on (a), we disable G$_1$ by replacing it with EG$_{1}$. Comparing (b) with (a) in Table~\ref{t_RG}, both of which carry out the guided convolution technology only once, although the error of (c) goes down a little bit, the GPU memory is heavily reduced by 11.95GB. These results give strong evidence that our new guidance design is not only effective but also efficient.

\begin{table*}[t]
\centering
\renewcommand\arraystretch{1.2}
\resizebox{0.74\textwidth}{!}{
\begin{tabular}{l|c|cccc|ccc|cc}
\hline
\multirow{2}{*}{Method}  & \multirow{2}{*}{RHN$_{3}$}  & \multicolumn{4}{c|}{RG} & \multicolumn{3}{c|}{AF}   & Memory        & RMSE  \\ \cline{3-9}
 & & G$_1$ & EG$_1$ & EG$_2$ & EG$_3$ & add  & concat & ours  & (GB)    & (mm) \\\hline\hline
baseline  &   &\checkmark&&&&&&& $\pm$0  & 778.6 \\ \hline
(a)  & \checkmark  &\checkmark&&&&&&& +1.35  &  769.0 \\ 
(b)         &\checkmark                       & & \checkmark  &            &            &            &              &          & -10.60   & 768.6 \\
(c)         &\checkmark                             &          & \checkmark  & \checkmark &            &            &              &          & +2.65    & 762.3 \\
(d)          &\checkmark                            &          & \checkmark  & \checkmark & \checkmark &            &              &          & +13.22   & 757.4 \\ \hline
(e)           & \checkmark                          &          & \checkmark  & \checkmark & \checkmark & \checkmark &              &          & +13.22   & 755.8 \\
(f)          &\checkmark                            &          & \checkmark  & \checkmark & \checkmark &            & \checkmark   &          & +13.22    & 754.6 \\
(g)          & \checkmark                            &          & \checkmark  & \checkmark & \checkmark &            &              &\checkmark& +13.28 & \textbf{752.1} \\ \hline
\end{tabular}
}
\caption{Ablation studies of RG/AF on KITTI validation set. RG-EG$_k$ refer to the case where we repeatedly use EG $k$ times. `$\pm0$' refers to 23.37GB. G$_1$ represents the raw guided convolution in GuideNet \cite{tang2020learning}, which is used only once in one fusion stage.}
\label{t_RG}
\end{table*}

\textbf{(ii) Repetitive guidance.} When the recursion number $k$ of EG increases, the errors of (c) and (d) are 6.3$mm$ and 11.2$mm$ significantly lower than that of (b) respectively. Meanwhile, as illustrated in Fig.~\ref{Fig.6}, since our repetition in depth (d) can continuously model high-frequency components, the intermediate depth feature possesses more detailed boundaries and the corresponding image guidance branch consistently has a high response nearby the regions. These facts forcefully demonstrate the effectiveness of our repetitive guidance design.

\textbf{(iii) Adaptive fusion.} Based on (d) that directly outputs the feature of RG-EG$_3$, we choose to utilize all features of RG-EG$_k$ ($k=1,2,3$) to produce better depth representations. (e), (f), and (g) refer to addition, concatenation, and our AF strategies, respectively. Specifically in (f), we conduct a $3 \times 3$ convolution to control the channel to be the same as RG-EG$_3$'s after concatenation. As we can see from the `AF' column of Table~\ref{t_RG}, all of the three strategies improve the performance of the model with a little bit GPU memory sacrifice (about 0-0.06GB), which demonstrates that aggregating multi-step features in repetitive procedure is effective. Furthermore, our AF mechanism obtains the best result among them, outperforming (d) 5.3$mm$. These facts prove that our AF design benefits the system better than simple fusion strategies. Detailed difference of intermediate features produced by our repetitive design is shown in Figs.~\ref{Fig.2} and~\ref{Fig.6}.

 \begin{figure*}[t]
  \centering
  \includegraphics[width=1\columnwidth]{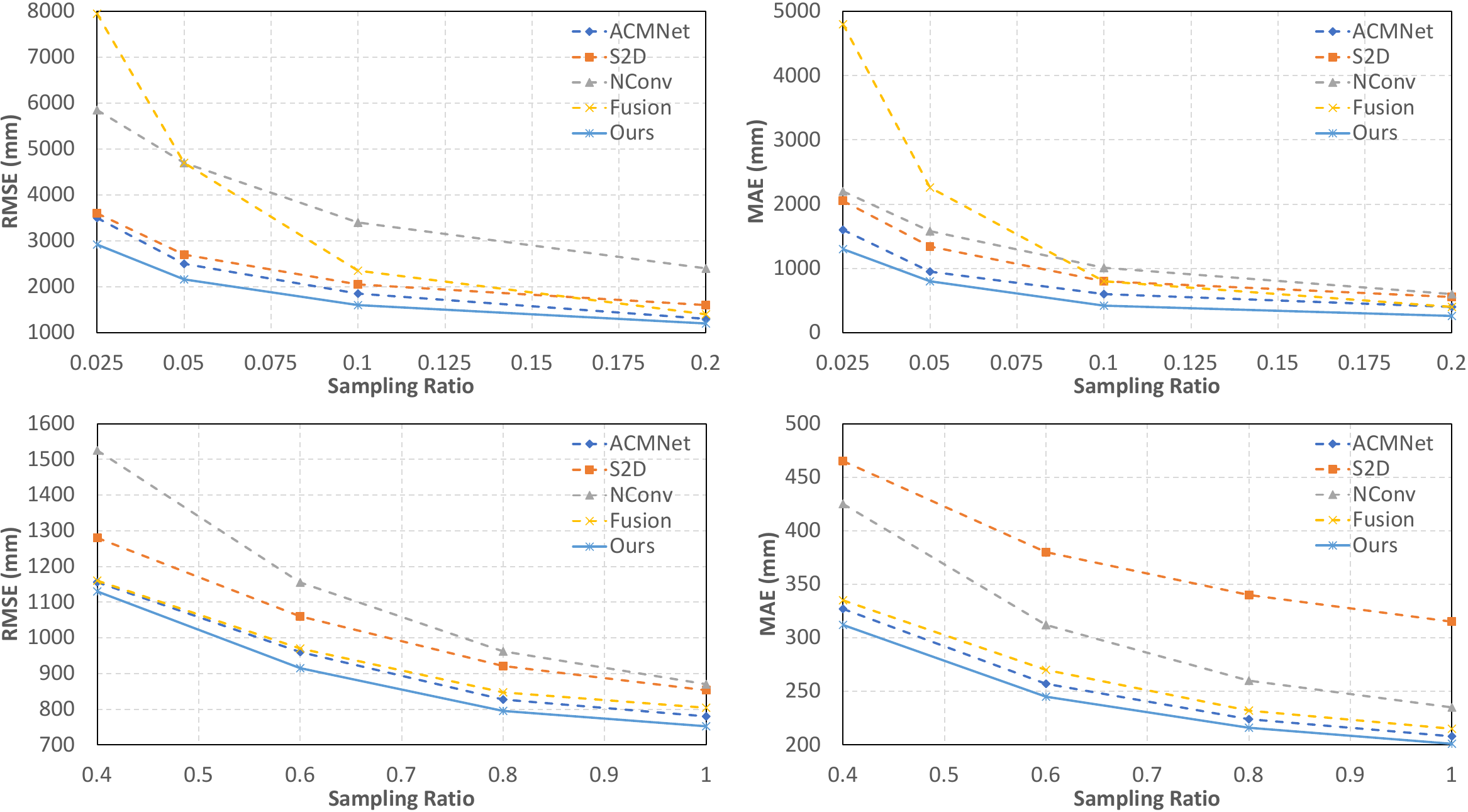}\\
  \caption{Comparisons under different levels of sparsity on KITTI validation split. The solid lines refer to our method while the dotted ones represent other approaches.}\label{density_kitti}
\end{figure*}

 \begin{figure*}[t]
  \centering
  \includegraphics[width=1\columnwidth]{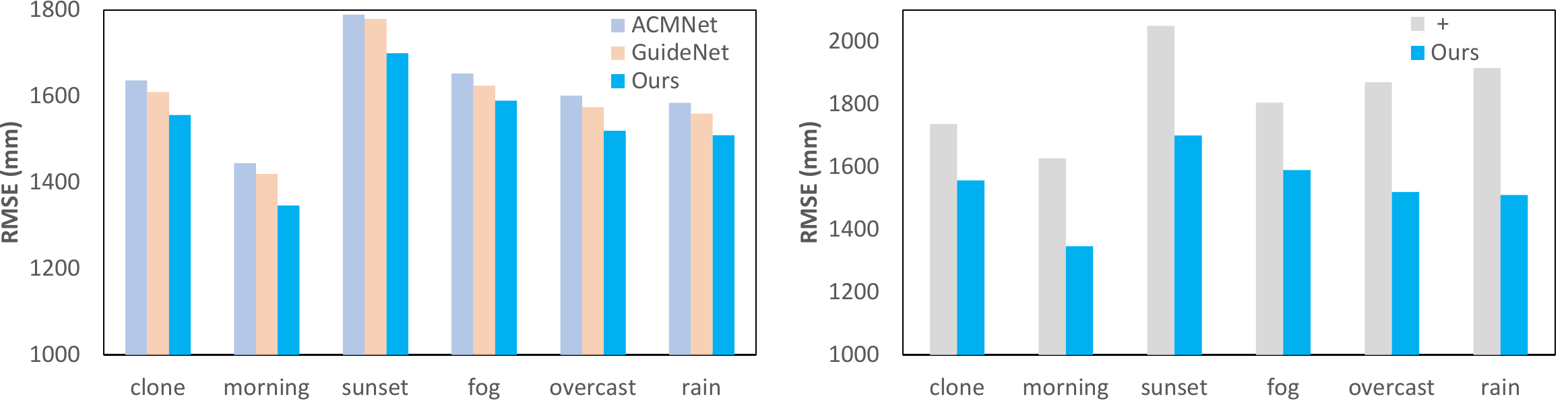}\\
  \caption{Comparisons with existing methods (left) and itself (right) replacing `RG' with `+', under different lighting and weather conditions on Virtual KITTI test split.}\label{Lighting_and_Weather}
\end{figure*}

\subsection{Generalization Capabilities}
In this subsection, we further validate the generalization capabilities of our RigNet on different sparsity, including the number of valid points, various lighting and weather conditions, and the synthetic pattern of sparse data. The corresponding results are illustrated in Figs. \ref{density_kitti} and \ref{Lighting_and_Weather}.

\noindent \textbf{(1) Number of valid points}

On KITTI selected validation split, we compare our method with four well-known approaches with available codes, \emph{i.e.}, S2D \cite{ma2018self}, Fusion \cite{vangansbeke2019}, NConv \cite{2020Confidence}, and ACMNet \cite{zhao2021adaptive}. Note that, all models are pretrained on KITTI training split with raw sparsity, which is equivalent to sampling ratios of 1.0, but not fine-tuned on the generated depth inputs. Specifically, we first uniformly sample the raw depth maps with ratios (0.025, 0.05, 0.1, 0.2) and (0.4, 0.6, 0.8, 1.0) to produce the sparse depth inputs. Then we test the pretrained models on the inputs. Fig. \ref{density_kitti} shows our RigNet significantly outperforms others under all levels of sparsity in terms of both RMSE and MAE metrics. These results indicates that our method can deal well with complex data inputs.

\noindent \textbf{(2) Lighting and weather condition}

The lighting condition of KITTI dataset is almost invariable and the weather condition is good. However, both lighting and weather conditions are vitally important for depth completion, especially for self-driving service. Therefore, we fine-tune our RigNet (trained on KITTI) on `clone' of Virtual KITTI \cite{gaidon2016virtual} and test under all other different lighting and weather conditions. As shown in the right of Fig.~\ref{Lighting_and_Weather}, we compare `RG' with `+' (replace RG with addition), 

\noindent our method outperforms `+' with large margin on RMSE. The left of Fig. \ref{Lighting_and_Weather} further demonstrates that RigNet has better performance than GuideNet \cite{tang2020learning} and ACMNet \cite{zhao2021adaptive} in complex environments. These results verify that our method is able to handle polytropic lighting and weather conditions.



In summary, all above-mentioned evidences demonstrate that the proposed approach has robust generalization capabilities.

\section{Conclusion}
In this paper, we explored the repetitive design in our image guided network for depth completion task. We pointed out that there were two issues impeding the performance of existing outstanding methods, \emph{i.e.}, the blurry guidance in image and unclear structure in depth. To tackle the former issue, in our image guidance branch, we presented a repetitive hourglass network to produce discriminative image features. To alleviate the latter issue, in our depth generation branch, we designed a repetitive guidance module to gradually predict structure-detailed depth maps. Meanwhile, to model high-frequency components and reduce GPU memory consumption of the module, we proposed an efficient guidance algorithm. Furthermore, we designed an adaptive fusion mechanism to automatically fuse multi-stage depth features for better predictions. Extensive experiments show that our method achieves outstanding performances.

\section{Acknowledgement}
The authors would like to thank reviewers for their detailed comments and instructive suggestions. This work was supported by the National Science Fund of China under Grant Nos. U1713208, 62072242 and Postdoctoral Innovative Talent Support Program of China under Grant BX20200168, 2020M681608. Note that the PCA Lab is associated with, Key Lab of Intelligent Perception and Systems for High-Dimensional Information of Ministry of Education, and Jiangsu Key Lab of Image and Video Understanding for Social Security, Nanjing University of Science and Technology.

\bibliographystyle{splncs04}
\bibliography{egbib}
\end{document}